# A General-purpose AI Avatar in Healthcare

**Nicholas Yan[1*], Gil Alterovitz[2],**

[1]Farragut High School, Knoxville, Tennessee, USA

[2]Biomedical Cybernetics Laboratory, Center for Biomedical Informatics, Harvard University, Boston, Massachusetts, USA

**\* Correspondence:**
Nicholas Yan
cubeymath@gmail.com

**Abstract**

Recent advancements in machine learning and natural language processing have led to the rapid development of artificial intelligence (AI) as a valuable tool in the healthcare industry. Using large language models (LLMs) as conversational agents or chatbots has the potential to assist doctors in diagnosing patients, detecting early symptoms of diseases, and providing health advice to patients. This paper focuses on the role of chatbots in healthcare and explores the use of avatars to make AI interactions more appealing to patients. A framework of a general-purpose AI avatar application is demonstrated by using a three-category prompt dictionary and prompt improvement mechanism. A two-phase approach is suggested to fine-tune a general-purpose AI language model and create different AI avatars to discuss medical issues with users. Prompt engineering enhances the chatbot's conversational abilities and personality traits, fostering a more human-like interaction with patients. Ultimately, the injection of personality into the chatbot could potentially increase patient engagement. Future directions for research include investigating ways to improve chatbots' understanding of context and ensuring the accuracy of their outputs through fine-tuning with specialized medical data sets.

## 1    Introduction

Recent developments in machine learning and natural language processing have allowed artificial intelligence (AI) to rapidly advance as a valuable tool in the healthcare industry. AI may run simulations of patient data in various scenarios, allowing them to assist doctors in their decision-making. AI can be utilized in various areas: detecting early symptoms of disease, assisting doctors in diagnosing patients, and conversing with patients. For example, IBM Watson for Oncology has been shown to diagnose breast cancer in accordance with human physicians 73% of the time (Jiang et al., 2017). Unlike humans, AI can easily keep up with the exponentially growing body of medical literature required to make an effective diagnosis. The ability of AI to infer a course of action from the ever-changing data in the present lets it outperform existing (e.g., treatment-as-usual) healthcare models (Bennett and Hauser, 2013). While AI applications in healthcare have been extensively studied, they are often trained for a single task using labeled, specific data, making deployment and generalization challenging. One open question calls for a general-purpose AI language model to perform diagnosis.

This paper will focus on the role of conversational AI, or chatbots, in healthcare. Chatbots use natural language processing to analyze users' queries and respond to them in a human-like manner (Jiang et al., 2017). Human doctors' diagnoses may not be completely accurate and doctors need time to prescribe the appropriate treatment. Chatbots can help mitigate this problem by serving as an intermediary between patient and doctor, giving adaptive and actionable health advice based on user input. As such, chatbots can benefit disadvantaged patients who may not have reliable access to health services while lessening the burden on healthcare workers who do not have to manage as many patients in-person.

Avatars may also make the usage of AI seem more appealing to prospective patients. In a study about the acceptability of AI in healthcare, one of the participants' major concerns was the lack of human connection (Nadarzynski, et al., 2019). When talking to a medical chatbot as a demo, they perceived the responses to be sterile and lifeless. On the other hand, patients who talked to computer-generated avatars expressed a greater level of trust and openness in interactions that would not have resulted otherwise. For instance, one patient who talked to a cube-like representation of their therapist in virtual reality reported that "it was easier to express [their] feelings" and "be [themselves] without being threatened." (Matsangidou, et al., 2022). After overcoming the initial hurdle- the skepticism about the new technology- the environment and character provided a sufficient disconnection from reality for the patient to share potentially useful information about their well-being.

The following research seeks to address the gap of AI avatars that are customizable in both knowledge and personality. This will employ a two-phase approach. Firstly, we fine-tune the general-purpose Generative Pre-trained Transformer 3.5 (GPT-3.5) by feeding a few examples of simulated patient cases through the ChatGPT user interface. The chatbot based on ChatGPT can respond to real patient requests properly with one- or few-shot prompting. The patient can then evaluate the AI model's diagnostic performance.

Secondly, we create different AI avatar or character profiles to discuss medical issues by using a three-category prompt dictionary. As stated previously, this will help users believe they are talking to a human and be more receptive to sharing personal details. Feeding sufficient examples to that model is required to adjust the model parameters. Fine tuning can not only improve the avatar and the AI-human interactions, but also enrich the medical knowledge of the avatar. With a wide variety of medical disciplines and possible personality traits (seen in the Methods), this framework can allow for the creation of hundreds of distinct prompt profiles, and by extension, avatars.

## 2   Background

As stated previously, conversational agents, or chatbots, use natural language processing to communicate and form relationships with users. Two well-known examples include Siri by Apple and Cortana by Microsoft. These chatbots introduced the concept of virtual assistants to the public while providing useful functionalities such as answering simple questions, scheduling tasks, and streamlining the user experience through integration with other apps. However, these older chatbots are constrained by their rule-based approach and limited understanding of context and lack of self-learning capabilities. Therefore, users may experience relatively rigid interactions and difficulty handling complex queries.

On the other hand, chatbots based on large language models (LLMs) benefit from vast pre-training on diverse datasets, allowing them to comprehend context, generate coherent responses, and exhibit human-like conversational abilities (Wei et al., 2023). LLMs typically rely on the transformer architecture. This model can perform self-attention, weighting the significance of different words in the text inputted and allowing it to better understand the contextual relationships between the words. With parallel processing, transformers can process all words in the input simultaneously, making them much more efficient (Vaswani et al., 2017).

This architecture enables LLMs to better remember the context of a user's messages throughout a conversation, reducing the chances of the user reminding the LLM of the topic. LLMs can also continually learn from user interactions to ensure a dynamic user experience (Jain et al., 2018). The utilization of deep learning techniques in LLMs has made the chatbot landscape more accessible to the general public while opening up further opportunities for chatbot applications. Indeed, several LLM-based healthcare chatbots have already been implemented for public use. Babylon Health, proposed by the UK National Health Service, is one example. The user shares their symptoms with the chatbot, which compares it with a database of cases treated by other doctors (Khadija et al., 2021).

Large language models have been widely celebrated for their remarkable language generation capabilities, but they come with inherent challenges. When presented with simple or ambiguous prompts, these models often produce broad and generic outputs. Prompt engineering can be used to combat this problem and create a chatbot that is convincingly human. A prompt is what the user inputs to guide the LLM to complete a certain task. Each output is influenced by the previous prompts, allowing the LLM to alter or refine its capabilities. For example, a prompt may instruct an LLM to respond in user-created syntax to make responses less verbose, behave as a certain entity to change the wording of its responses, or provide suggestions for future prompts (White et al., 2023). This approach allows users to achieve more accurate and tailored outputs, ensuring that the generated content aligns precisely with the intended context and requirements.

Prompt patterns are at the core of prompt engineering, as these have been shown to successfully systematically alter LLM interactions (White et al., 2023). Several relevant prompt patterns are introduced as follows:

> Audience Persona: The user assumes a persona (job position, historical figure, inanimate object, etc.) and the LLM gives a response tailored to the persona's knowledge. *(Example: Explain how to compose in a digital audio workstation for me. Assume I am Ludwig van Beethoven.)*
>
> Question Refinement: The user asks the LLM a question within a certain scope and provides details that let the LLM improve the user's prompt. *(Example: When I ask a question about shopping, suggest a better version that accounts for healthy eating and frugal spending instead.)*
>
> Fact Check List: The LLM adds a list of facts, true or not, to the end of the output, which can then be verified by the user. *(Example: From now on, append a set of facts about architecture that are contained in the output. The facts must be relevant to the output and be able to be checked for accuracy.)* (White et al., 2023)

Each pattern allows the LLM to create interesting and creative outputs that would not normally be considered otherwise. Using prompt patterns may lead to LLMs creating more targeted prompts themselves. Therefore, prompt patterns hold immense potential in enhancing the usability of LLMs and can be used to solve a variety of problems from gamifying tasks to troubleshooting code.

## 3  Methods

The healthcare AI avatar application is composed of a chatbot and an avatar. After starting the application, the patient selects an avatar of a certain medical specialty. Then, the patient picks the avatar's special characteristics or its personality. The application creates the prompt profile from the prompt dictionary from the user's selections. In summary, this profile covers the avatar's medical knowledge, common characteristics as a good doctor and special characteristics involved doctor's personality. ChatGPT has been used as the ANN (artificial neural network) due to its flexibility, ease of use, and ubiquity. A sample of the prompt dictionary which only includes the medical knowledge has been reproduced below, while the aforementioned steps have been summarized in Figure 1.

>   Avatar role: General practice
>   Prompt words: In this dialogue session with me, you are a doctor with a specialty in general practice. You have the following medical knowledge: comprehensive patient care and management; broad medical knowledge spanning various disciplines; ability to diagnose and treat a wide range of acute and chronic conditions; emphasizing preventive medicine and health promotion; continuity of patient care over time; strong communication skills to build patient-doctor relationships.

>   Avatar role: Allergy
>   Prompt words: In this dialogue session with me, you are a doctor with a specialty in allergies. You have the following medical knowledge: expertise in diagnosing and managing allergies and immunologic disorders; knowledge of allergens and their effects on the body; skill in administering and interpreting allergy tests; ability to develop treatment plans, including immunotherapy if necessary.

**Figure 1**
*Sample chatbot workflow featuring three-category prompt profile*

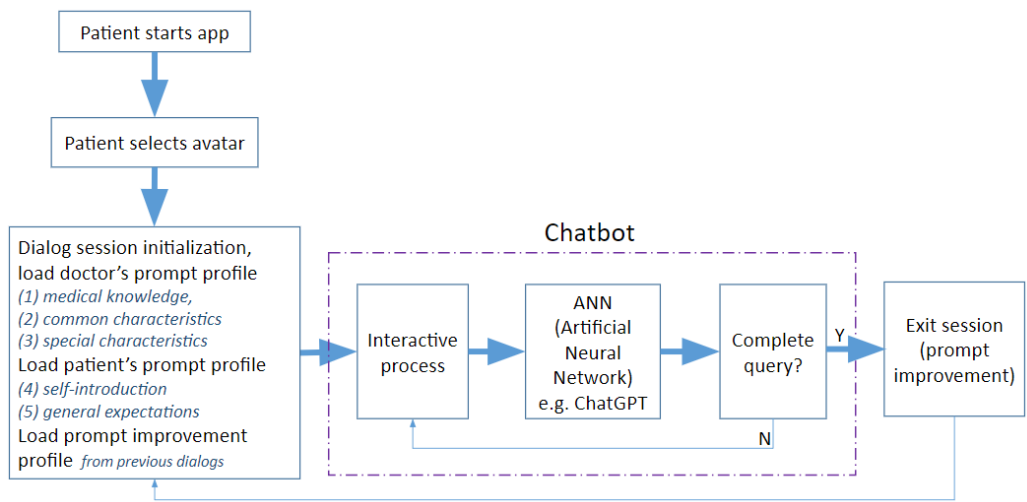

ChatGPT takes the prompt profiles of the patient and the doctor and engages the patient in the interactive process. From this point, ChatGPT acts as a chatbot and continues the conversation based on the submitted prompt profile. The patient can choose to close this dialogue session at any step while the application can collect all the responses from the user and ChatGPT in the dialogue for prompt improvement. The improved prompt can be applied to the next dialogue when the patient restarts the application.

We begin by creating a prompt profile for each chatbot from the three-category prompt dictionary. Each prompt is composed of three parts:

> Medical knowledge (given by the chatbot's specialty)
> Common characteristics
> Selectable special characteristics

The common characteristics of each chatbot are the agreed-upon characteristics of exceptional doctors. According to a survey of adult patients, exceptional doctors impress patients with their overall qualities and excel in communication skills, trust-building, and patient safety (Schnelle and Jones, 2023). They empower patients by involving them in decision-making and treatment processes and exhibit expertise in multiple areas. On the other hand, the special characteristics are defined as the chatbot's personality traits which can be added to the prompt to give each chatbot a more human feel and to differentiate one chatbot from another. These were derived from Patrick Gunkel's categorization of over 600 personality traits and consolidated into the following categories. Analogous traits (e.g., authentic, sincere) were combined for brevity, though using either would lead to different outputs from the LLM and should be considered.

> Social traits: Kind, empathetic, sociable
> Emotional traits: Optimistic, resilient, serene
> Dynamic traits: Courageous, enthusiastic, adaptable
> Ethical traits: Virtuous, fair, trustworthy
> Positive traits: Positive, humorous, joyful
> Organizational traits: Orderly, punctual, thorough
> Leadership traits: Responsible, inspirational, decisive
> Mindful traits: Contemplative, thoughtful, aware
> Intellectually honest traits: Authentic, sincere, transparent (Gunkel, 2006)

## 4    Results

The workflow concept in the ChatGPT environment is shown in Figure 2. After the user begins the chat, they can select any avatar from the prompt dictionary. In this example, the user selects the general practitioner. The common characteristics give ChatGPT the qualities of an exemplary doctor and the special characteristics give it human-like personality traits, in this case trustworthiness. The combined prompt, following the template in the Methods section, is used as the avatar prompt profile for the initialization of this dialogue session. If ChatGPT's response is deemed unsatisfactory, the user can regenerate the response and send feedback through ChatGPT's built-in features.

A patient's case from the American College of Radiology has been selected as an example to compare the outputs of generic ChatGPT with the specialized chatbot as seen in Figure 2 (Rao et al., 2023). Both outputs use zero-shot prompting; in other words, the LLM is given no examples to learn from before engaging in dialogue from the user. The output on the left is by generic ChatGPT while the output on the right is by the specialized chatbot avatar.

**Figure 2**
*Comparison of ChatGPT's responses with a generic prompt versus a specialized prompt. (A) ChatGPT output after only using the patient case as input. (B) Refined prompt using the three-category prompt dictionary. (C) ChatGPT output after using the refined prompt and the patient case as input.*

Patient age: 24 years old
Sex: Female
Chief Complaint: Complaints in the lower extremity during activity. The patient presents with complaints of pain in the left lower extremity during physical activity, which has been progressing over the past few months. The patient is an avid runner, and she has been training for a marathon for the past year. She has no significant medical history and takes no medications. On physical examination, the patient has normal vital signs. There is no edema in the lower extremities. The left lower extremity appears to be slightly smaller than the right one, and there is mild tenderness in the popliteal fossa.

Given the following case above, please diagnose the patient.

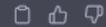
As an AI language model, I don't have the capability to diagnose medical conditions, as it requires a proper medical examination, laboratory tests, and often imaging studies by a qualified healthcare professional. My purpose is to provide general information and support, but not to offer medical diagnoses.

Based on the information provided, it seems the patient is experiencing pain in the left lower extremity during physical activity, along with some size discrepancy and tenderness in the popliteal fossa. There are several potential causes for such symptoms, and the most appropriate action would be to consult a healthcare professional, such as an orthopedic specialist or sports medicine physician.

Some possible differential diagnoses for lower extremity pain during activity could include:

1. **Overuse injuries:** Running frequently and training for a marathon can lead to overuse injuries, such as stress fractures, tendinitis, or muscle strains.
2. **Popliteal artery entrapment syndrome:** A condition where the popliteal artery is compressed during physical activity, causing pain and reduced blood flow to the leg.
3. **Baker's cyst:** A fluid-filled cyst that can form in the popliteal fossa due to joint inflammation or injury.
4. **Deep vein thrombosis (DVT):** A blood clot that forms in a deep vein, causing pain and swelling.
5. **Peripheral artery disease (PAD):** Reduced blood flow to the legs due to narrowed arteries.
6. **Nerve compression:** Compression of nerves in the lower extremity, such as the sciatic nerve.

It's essential for the patient to schedule an appointment with a healthcare professional for a thorough evaluation, physical examination, and potentially further investigations like imaging or blood tests to determine the exact cause of the symptoms and develop an appropriate treatment plan. Self-diagnosis and treatment are not advisable in this case.

(A)

(B)

I

Generic ChatGPT provides a broad, unfocused list of possible diagnoses for the patient. On the other hand, the chatbot that acts as a general practitioner hones in on a specific diagnosis- namely, popliteal artery entrapment syndrome- and outlines specific details from the patient's report to support the diagnosis. Alternate treatments similar to generic ChatGPT have not only been provided but also expanded upon (for example, the general practitioner chatbot suggests additional tools such as ultrasound and MRI for further evaluation).

The chatbot's assumed personality traits are most discernable immediately after the user enters the desired prompt profile ("In this dialogue session with me…"). Both the prompt profile and chatbot output follow the same structure; the specialty knowledge is discussed first, followed by the chatbot's general and special characteristics. Notably, the chatbot tends to use identical words and phrases found in the prompt profile in its first output, especially in terms of personality traits. After the patient's case is given, however, the chatbot assumes a more professional demeanor.

## 5 Discussion

### 5.1 Implications

This work demonstrates a framework for the general-purpose AI avatar application in healthcare. Recent LLM versions such as ChatGPT 3.5 perform well with the prompt scheme of a pre-configured patient prompt profile and the doctor's prompt profile. Even though the prompt in Figure 2b did not contain any specific medical information, it still allowed ChatGPT to output a more specific and detailed diagnosis compared to generic ChatGPT. In short, it is the ability to delve into a specific medical problem and propose relevant diagnostic tests that emphasizes the potential practical utility of specialized chatbots.

Furthermore, this method shows the potential of using prompt engineering to let ChatGPT interact with the patient in a conversational manner and improve patient engagement. In addition, it emphasizes the important role that the chatbot's personality plays in human-chatbot interactions. Users presented with chatbots of different specialties expect the chatbot's specialty to match its personality; for example, a medical chatbot is expected to speak in a formal tone. The significant impact of a chatbot's personality on a user is evident when they choose to refer to the chatbot using he/she pronouns, deviating from the customary "it" used for inanimate objects or tools (Jain et al., 2018). As such, the creative responses from the chatbot combined with the patient's personal connection to the chatbot may further engage patients and prolong their conversations, making it more likely for patients to provide the chatbot with pertinent information.

### 5.2 Limitations and Future Directions

LLMs have flaws that persist even with careful prompt engineering. One major concern is the chatbot's potential for misinformation or lack of information. Firstly, LLMs such as ChatGPT have knowledge cutoff dates; consequently, the model might unknowingly provide outdated or inaccurate information to users. Secondly, the training process of these language models relies on diverse datasets from the internet instead of specialized medical sources, potentially propagating false information from unverified sources (Ray, 2023). Moreover, the reliance on internet-derived data poses an inherent risk of incorporating biased content, inadvertently amplifying existing societal biases (Ray, 2023). LLMs may also oversimplify information in its diagnoses, leading to omission of information and incorrect conclusions. Rarer conditions appear less often in datasets by nature, skewing the LLM to make inaccurate conclusions regarding these conditions (Elkassem and Smith, 2023). Thus, patients should not overly rely on LLMs and should consider their outputs in tandem with human advice to ensure their safety.

Future research may further investigate human-chatbot interactions in order to further improve the chatbots' prompt engineering. For instance, one major aspect to overcome is the chatbot's limited understanding of context within and between user sessions. Participants who used a shopping chatbot found themselves disappointed when the chatbot did not remember conversational information exchanged during previous sessions, such as the item recommendations (Jain, et al., 2018). In a medical setting, it would be similarly frustrating for a patient to have to prime the chatbot with the same initial information for every session. Due to the random nature of LLM outputs, it is also unlikely for the chatbot to reach the fixed stage in the conversation that the patient wants. If a conversation becomes long enough, this problem is inevitable due to the LLM's finite context window- the LLM's memory that stores user prompts and its responses. However, future models will likely have wider context windows to account for longer conversations. In addition, the proposed prompt improvement concept can consolidate information to make conversations more efficient (use less tokens) and allow for the conversations in one dialogue session to be collected and sent to ChatGPT for future sessions.

The accuracy of the chatbots' outputs against those of professionals and non-professionals are being actively studied across various disciplines (Levine et al., 2023; Singhal et al., 2023; Bakhshandeh, 2023; Pataranutaporn et al., 2021). However, the method of generating multiple responses allows for patients and doctors to cross-reference information between responses and find commonalities between them. Human doctors can then act as quality control to provide specialized or up-to-date advice that the chatbot does not have. Furthermore, the LLM can be fine-tuned using data sets (e.g., medical texts, case studies) specifically tailored towards the field of interest. Projects such as BioGPT, PubMedGPT, and Galactica already show the potential of training LLMs on large bodies of medical text (Singhal et al., 2023). For this project, however, this must be repeated multiple times for each field.

Additionally, the ethics of LLM-based chatbots, especially with regards to user safety and privacy, are being actively researched. One issue is allowing users to delete their data. Even though LLMs have limited memory due to the lengths of their context windows, they do not "forget" like humans do. Instead, neural networks incorporate new data points by adjusting their weights and parameters as a whole (Hinton, 1992). This would make it near-impossible to trace a singular point of data that a patient may want to remove from the AI's knowledge base. Implementing measures for LLMs to methodically "unlearn" data while maintaining its previous performance should be considered in future studies.

## 6   Conclusion

In this paper, we explored the usage of a three-category prompt system for an LLM-powered medical chatbot to improve patient engagement. The demonstration of the workflow confirms the success of the prompt system, as the chatbot has taken on the common and special characteristics as outlined in the prompt. The personification of the chatbot will promote more meaningful conversations with users as the users interact with the chatbot more. Due to current LLM limitations, users should be cognizant of the limits of the information that chatbots output and should use that information in conjunction with advice from a human doctor. However, embedding specialized medical knowledge in the chatbot combined with improvements in future models (such as larger context windows) may improve the reliability and friendliness of the chatbot.

## 7   Conflict of Interest

The authors declare that the research was conducted in the absence of any commercial or financial relationships that could be construed as a potential conflict of interest.

## 8   Author Contributions

## 9   Funding

The funding of this study was provided by the Research Science Institute and the Center for Excellence in Education.

## 10    Acknowledgments

I would like to deeply thank Dr. Gil Alterovitz and Ning Xie for their mentorship and streamlining the research process; Shuvom Sadhuka, Michael Huang, and Dr. Steve Fein for their valuable feedback and encouragement as this paper developed.

## 12   Data Availability Statement

The original contributions presented in the study are included in the article. Further inquiries can be directed to the corresponding author.